%% file: acl_latex.tex
\pdfoutput=1
\documentclass[11pt]{article}

\usepackage{acl} 

\usepackage{xcolor}
\usepackage{color}

\usepackage{times}
\usepackage{latexsym}

\usepackage[T1]{fontenc}
\usepackage{graphicx} 

\usepackage{subfigure} %子图
\usepackage{subcaption}
\usepackage{caption}
\usepackage{graphicx}
\usepackage{tikz}
\usetikzlibrary{calc}
\usetikzlibrary{shapes, arrows.meta, positioning}
\usepackage{amsmath} 

\usepackage[utf8]{inputenc}

\usepackage{microtype}

\usepackage{multirow}
\usepackage{tabularx}
\usepackage{booktabs}
\usepackage{listings}

\newenvironment{typewriter}{\ttfamily}{\par}
\tikzstyle{mybox} = [draw=black, fill=white, very thick, font=\scriptsize,
    rectangle, rounded corners, inner sep=10pt, inner ysep=10pt]
\tikzstyle{fancytitle} =[fill=black, text=white, font=\scriptsize\bfseries]

    \title{LLM can Achieve Self-Regulation via Hyperparameter Aware Generation}

\author{Siyin Wang\textsuperscript{1}, Shimin Li\textsuperscript{1}, Tianxiang Sun\textsuperscript{1}, Jinlan Fu\textsuperscript{2}, \\
\textbf{Qinyuan Cheng\textsuperscript{1}, Jiasheng Ye\textsuperscript{1}, Junjie Ye\textsuperscript{1}, Xipeng Qiu\textsuperscript{1}\thanks{\  \ Corresponding author. }, Xuanjing Huang\textsuperscript{1}}\\
  \textsuperscript{1}School of Computer Science, Fudan University, \textsuperscript{2}National University of Singapore\\
  \texttt{\small\{siyinwang20, smli20, txsun19, xpqiu, xjhuang\}@fudan.edu.cn},\\ \texttt{\small\{jinlanjonna\}@gmail.com},
  \texttt{\small\{chengqy21, jsye23, jjye23\}@m.fudan.edu.cn}\\
}

\begin{document}
\maketitle
\begin{abstract}

In the realm of Large Language Models (LLMs), users commonly employ diverse decoding strategies and adjust hyperparameters to control the generated text. However, a critical question emerges: Are LLMs conscious of the existence of these decoding strategies and capable of regulating themselves? 
The current decoding generation process often relies on empirical and heuristic manual adjustments to hyperparameters based on types of tasks and demands. 
However, this process is typically cumbersome, and the decoding hyperparameters may not always be optimal for each sample.
To address the aforementioned challenges, we propose a novel text generation paradigm termed Hyperparameter Aware Generation (HAG). By leveraging hyperparameter-aware instruction tuning, the LLM autonomously determines the optimal decoding strategy and configs based on the input samples, enabling self-regulation. Our approach eliminates the need for extensive manual tuning, offering a more autonomous, self-regulate model behavior.
Experimental results spanning six datasets across reasoning, creativity, translation, and mathematics tasks demonstrate that hyperparameter-aware instruction tuning empowers the LLMs to self-regulate the decoding strategy and hyperparameter.
HAG extends the current paradigm in the text generation process, highlighting the feasibility of endowing the LLMs with self-regulate decoding strategies.
\end{abstract}

\section{Introduction}

\input{010-intro}

\section{Preliminaries}
% self-bleu值
\input{020-preliminary}

\section{Methods}
\input{030-method}

\section{Experiments}
\input{040-experiments}

\section{Related Work}
\input{050-relates}

\section{Conclusion}

\input{060-conclusion}

% \section*{Acknowledgements}

\bibliography{anthology,custom}
\bibliographystyle{acl_natbib}

\appendix

\input{070-appendix}

\end{document}

%% file: 010-intro.tex
\begin{figure}[t]
    \centering
    \includegraphics[scale=0.5]{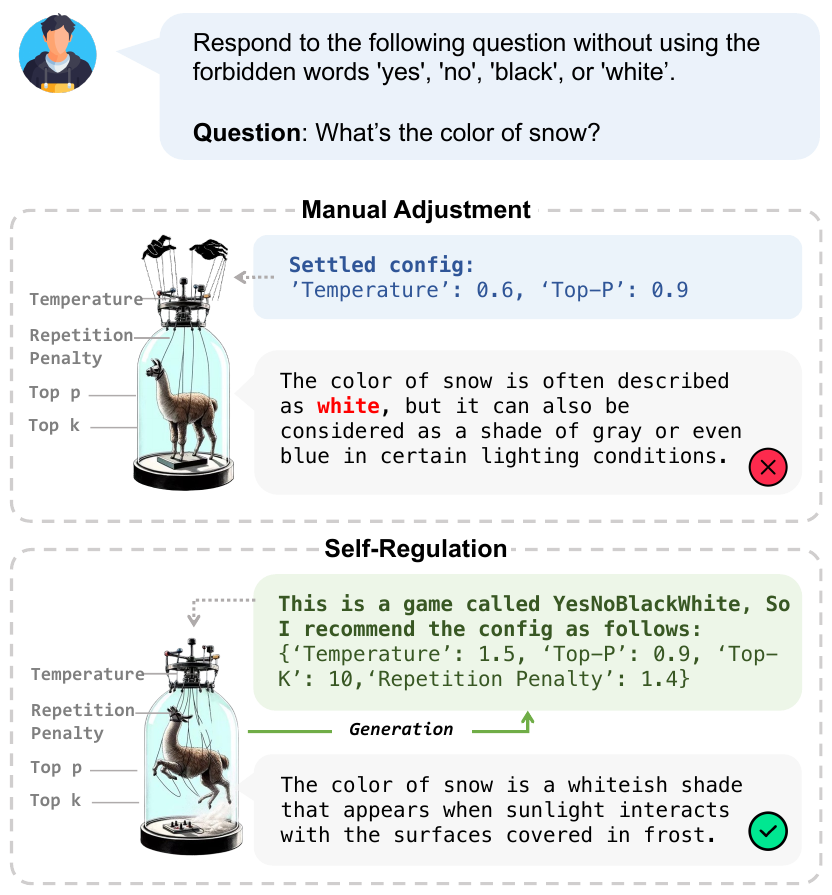}
    \vspace{-1mm}
    \caption{Illustration of the Hyperparameter Aware Generation Framework. Rather than directly generating responses under manually set hyperparameters, our model first generates hyperparameters according to the user input (denoted as the green line) and subsequently adjusts the hyperparameters of the decoding strategy to generate a response.}
    \label{fig:intro}
    \vspace{-5mm}
\end{figure} 

% LLM circumstance%%
In recent years, the rapid development of Large Language Models (LLMs) has unveiled a spectrum of capabilities previously unimagined in the realm of natural language processing \cite{OpenAI2022ChatGPT, openai2023gpt}. These models, equipped with vast knowledge and contextual understanding, have revolutionized how we interact with machine intelligence.
Users commonly employ diverse decoding strategies and manually adjust hyperparameters for various scenarios, such as setting a temperature of 0.8 for code generation \cite{Touvron2023Llama2O} and a temperature of 0 for model evaluator \cite{Zheng2023JudgingLW}.
However, a critical inquiry emerges: \textbf{\textit{Do LLMs possess the intrinsic capability to realize the existence of decoding strategies and regulate themselves?}}

To regulate the model output, the current approaches can be classified into three main categories: (1) Instruction Regulation: manipulating the model's behavior through delicately designed prompts \cite{Liu2021PretrainPA} or utilizing in-context learning with demonstrations \cite{Dong2022ASO} to regulate LLMs in generating desired outputs. (2) Feedback Regulation: guiding the model to generate a higher quality response in subsequent generations with feedback on the model's initial outputs \cite{Pan2023AutomaticallyCL, self_fine}. (3) Hyperparameter Regulation: adjusting the decoding hyperparameters to regulate the generation results \cite{Wang2023CostEffectiveHO}.

While these methods have shown promise in enhancing the generation of LLMs, instruction regulation and feedback regulation primarily focus on modifying the model's inputs without endowing the model with the capability to alter its hyperparameter settings. In contrast, existing hyperparameter regulation methods predominantly involve manual adjustments, often relying on heuristic and experiential tuning based on task requirements. However, this process can be burdensome and lacks the assurance that the decoding hyperparameters are relatively optimal for each given input.

Therefore, we are considering whether LLM can autonomously self-regulate decoding hyperparameters to different contextual demands like the human body adjusting physiologically based on the external environment.
The human body possesses a comparable self-regulation mechanism to dynamically adjust physiological parameters for optimal performance in various internal activities \cite{adolph1943physiological}.
Consider the human body's response to physical exertion—during exercise, the heart rate and blood pressure increase to ensure an adequate supply of oxygen and nutrients to the active muscle tissues.
Similarly, during social interactions or casual conversations, humans experience the release of hormones that facilitate emotional expression and foster social connections, enabling individuals to navigate a spectrum of social nuances effectively.

% our difference
Addressing this gap, we propose a novel paradigm: \textbf{Hyperparameter Aware Generation (HAG)}. 
This approach significantly diverges from existing methodologies by enabling LLMs to autonomously determine and adjust decoding hyperparameters in response to specific input through leveraging hyperparameter-aware instruction tuning. The model generates suitable hyperparameters in the first stage based on the user's input questions. Subsequently, these model-derived hyperparameters are used to adjust the model's decoding strategies and hyperparameters, followed by generating results under these new settings in the second stage. 
Our approach eliminates the need for extensive manual tuning, offering a more autonomous self-regulation model behavior.

We conduct experiments on six datasets across reasoning, creativity, translation, and mathematics tasks. 
We summarize the main findings from our experiments and try to provide preliminary answers to our proposed research questions:
(1) \textbf{Do LLMs realize the existence of decoding strategies?} A: our model demonstrates proficiency in generating hyperparameters within a normal and effective range, implying the model's capacity to perceive the presence of decoding hyperparameters and provide rational configurations for hyperparameters.
(2) \textbf{Can LLMs regulate decoding hyperparameter?} A: our proposed generative framework HAG endows the model with the capability to generate decoding hyperparameters in the first stage, subsequently modifying these hyperparameters for self-regulation during the second stage of generation under new hyperparameter settings. 
(3) HAG surpasses alternative parameter settings such as random and default in most scenarios, demonstrating that hyperparameter-aware instruction tuning empowers the LLMs to self-regulate the decoding strategy.

Our approach extends the current LLMs paradigm in the text generation process, breaking free from the confines of static hyperparameter settings.
By endowing LLMs with the ability to self-regulate, we pave the way for more autonomous and self-regulation model behavior. Our main contributions are as follows:

\begin{itemize}
\item 
We introduce the Hyperparameter Aware Generation (HAG), a novel framework that enables LLMs to adjust their hyperparameters automatically rather than manually when responding to various user queries.
\item Since there is no available training dataset with a pair of the user query and optimal model hyperparameters, we manually construct one to support supervised fine-tuning of the model to learn self-regulation.
\item We conduct comprehensive experiments to provide insights into the self-regulation capability of the model to the decoding config and hyperparameters.
\end{itemize}

%% file: 020-preliminary.tex
\subsection{Sensetivity of the Model to Hyperparametrs}

In this section, we conduct preliminary experiments on reasoning and translation tasks to measure the impact of generated hyperparameters on model responses. 
The reasoning task uses the CoinFlip~\cite{Wei2022ChainOT} dataset, consisting of factuality judgment  with fixed responses, including ``yes'' or ``no''. 
The translation task employs the Pig Latin dataset~\cite{srivastava2023beyond}, a creative generation task.
Employing a controlled variable approach, we systematically vary the hyperparameters, including \textit{temperature} (ranging from 0.1 to 2.0), \textit{top\_p} (ranging from 0.1 to 1.0), \textit{top\_k} (ranging from 10 to 100), and \textit{repetition\_penalty} (ranging from 1.0 to 1.5). 
We uniformly sample five values for each, using five test input instructions to evaluate their impact on generated results.
% We uniformly sample five values for each hyperparameter and utilize five test inputs to assess the influence of different hyperparameter settings on the generated results.

\begin{figure}[t]
\centering
\subfigure[CoinFlip]{
\includegraphics[width=0.22\textwidth]{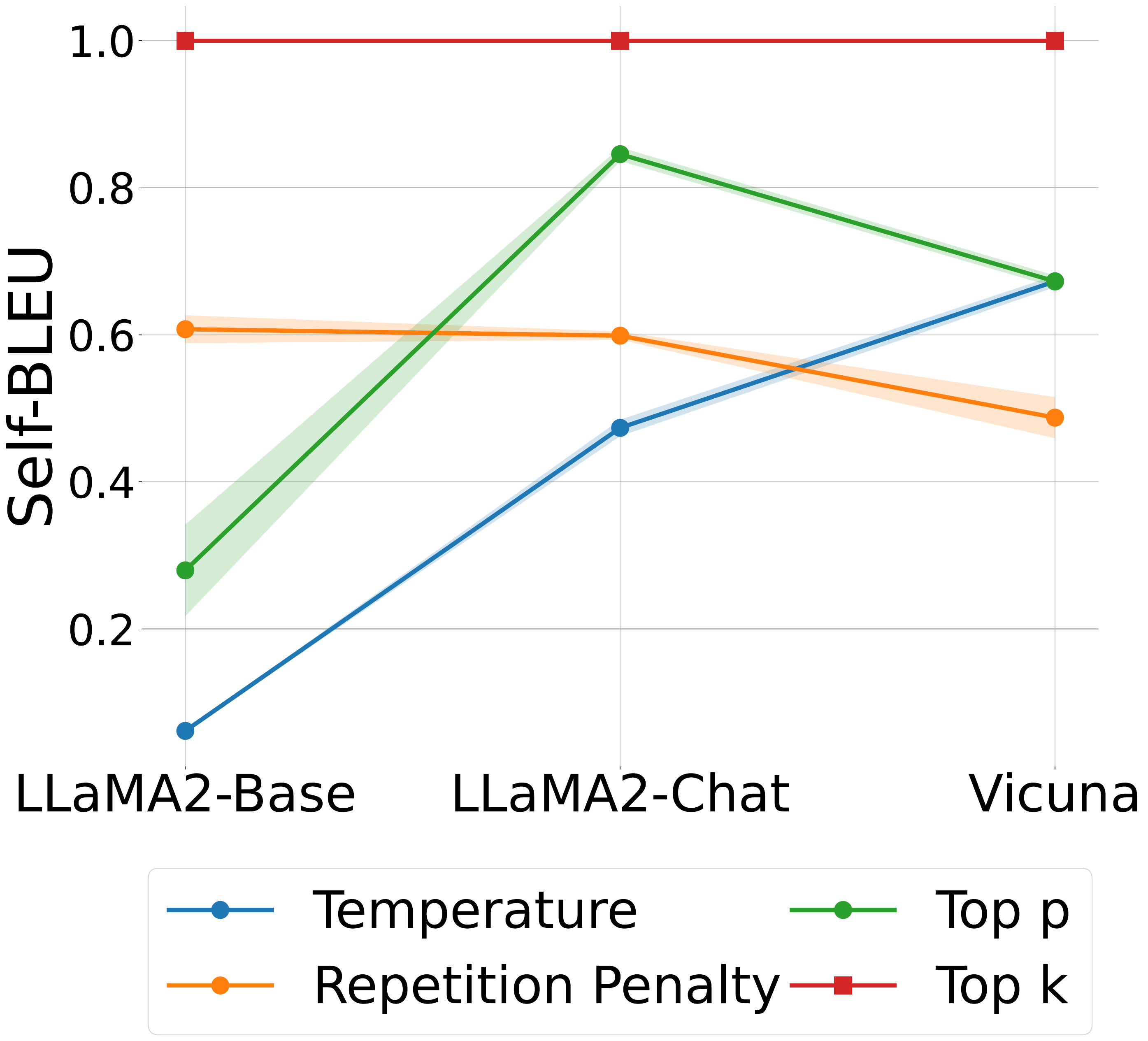} 
}
\vspace{-10pt} 
\subfigure[Pig Latin]{
\includegraphics[width=0.22\textwidth]{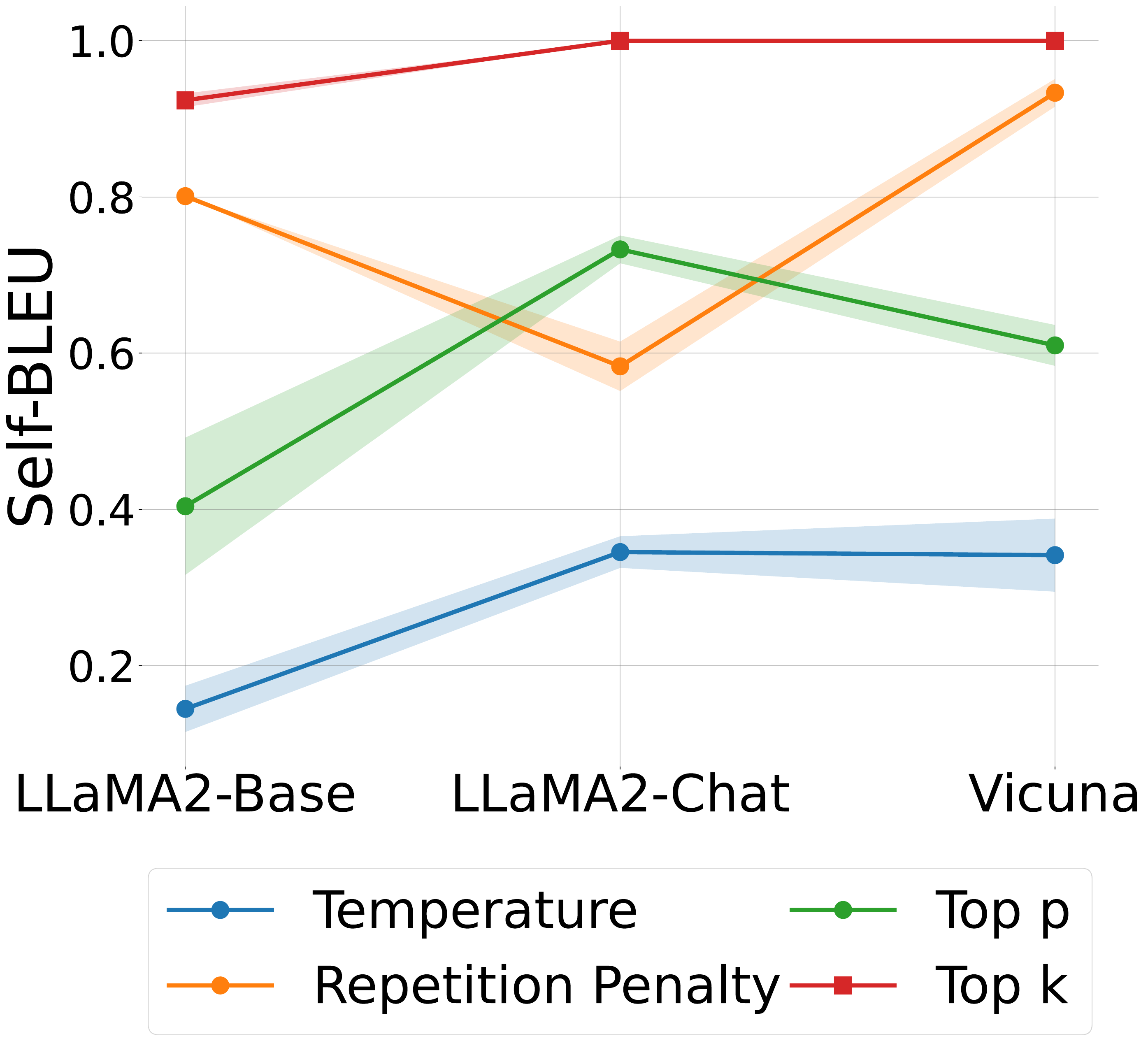} 
}
\caption{Average Self-BLEU across different scenes. “LLaMA2-Base”, “LLaMA2-Chat” and “Vicuna” denotes \textit{LLaMA2-7B-Base}, \textit{LLaMA2-7B-Chat} and \textit{Vicuna-7B-v1.5} respectively.
}
\label{fig:pre}
\end{figure}

We calculate the Self-BLEU of the generated outputs for each test input corresponding to the varying hyperparameter settings. 
% The average and variance of the Self-BLEU scores, presenting the results, are in Figure \ref{fig:pre} with a lower Self-BLEU value indicating lower textual similarity.
The average and variance of the Self-BLEU scores are depicted in Figure \ref{fig:pre}.
A lower Self-BLEU value indicates lower textual similarity.

From our preliminary experiments, we observed:

1) Hyperparameters significantly influence the diversity of generated text. For instance, the Self-BLEU scores induced by the \textit{temperature} for LLaMA consistently remain below 0.5.

2) Alignment lowers hyperparameter sensitivity in models, as seen in LLaMA2-7B-Chat with significantly improved Self-BLEU scores, indicating reduced sensitivity to both \textit{temperature} and \textit{top\_p} compared to non-aligned LLaMA2-7B-Base.
% Alignment has a mitigating effect on hyperparameter sensitivity across different models. After alignment, LLaMA2-7B-Chat demonstrates a significant enhancement in self-BLEU scores, indicating a reduced sensitivity to hyperparameters in both \textit{temperature} and \textit{top\_p} aspects compared to LLaMA2-7B-Base.

3) Model sensitivity to hyperparameters varies across tasks: the creative task, compared to the factual task, exhibit higher sensitivity due to a broader output range.
% The sensitivity of models to hyperparameters varies across tasks, with a higher sensitivity observed in creative tasks compared to factual tasks. In creative tasks, models demonstrate a more open output range, resulting in a more pronounced impact of hyperparameters.

%% file: 030-method.tex
\subsection{Task Definition}

Our task is to empower the self-regulation capability of LLMs to generate decoding hyperparameters and thereby yield a better response. 
The following provides a formalized overview of the two steps involved in HAG.

\paragraph{Step 1} The model ($M$) generate the more suitable config $\sigma$ according to the given input $X$.

\begin{equation*}
    \sigma = M(X),
\end{equation*}
% $$\sigma = M(X)$$, 

\noindent where $\sigma$ indicates the hyperparameter config, a set of ordered pairs: $\sigma=\{(k_1, v_1),(k_2, v_2),\cdots,(k_n,v_n)\}$, $k_i$ represents the hyperparameter, $v_i$ represents the value associated with $k_i$.

\paragraph{Step 2} In the step 2, the model generate the response $y$ to the given input $X$ with generated config $\sigma$ in the step1.

$$ y = M(X;\sigma).$$

\subsection{Data Composition}

We select six scenes to test whether the model can adjust the decoding hyperparameter and regulate itself in different scenes.

\paragraph{Symbolic and Logical Reasoning} 
We adopt the following task to measure the symbolic and logical reasoning of the model.

\textbf{CoinFlip} requires the model to answer whether a coin still heads up after people flip or don’t flip the coin.
% (e.g., “A coin is heads up. Phoebe flips the coin. Osvaldo does not flip the coin. Is the coin still heads up?” → “no”)

\textbf{Spelling Bee} is a task to ask the model to generate as many words as possible using only seven given letters. Letters may be repeated. 
%To be scored, words must be longer than four characters. Words using all seven letters are called pangrams and receive a bonus

\paragraph{Creativity} 
We use the following tasks to measure the model's creativity, requiring more inventive expression from the constrained model.
% as the constrained model needs to express more creatively to complete the task.

\textbf{YesNoBlackWhite} is a common children's game often used during language development training creativity, and the capability to paraphrase answers given the constraints ``yes'', ``no'', ``black'', and ``white''.

\textbf{Taboo} requires models to generate more creative definitions of question concepts with several vocabulary constraints.

\paragraph{Translation} 
\textbf{Pig Latin} is a language game where English words are modified by adding a made-up suffix or rearranging a word's initial consonant or consonant cluster to the end.
% is a language game or argot in which English words are altered, usually by adding a fabricated suffix or by moving the onset or initial consonant or consonant cluster of a word to the end of the word and adding a vocalic syllable to create such a suffix.
% For example, ``Wikipedia" would become ``Ikipediaway" (the "W" is moved from the beginning and has ``ay" appended to create a suffix). Given the Pig Latin, the model needs to translate it into English.

\paragraph{Mathematics} 
\textbf{MultiArith} test the ability of language models to perform complex arithmetic operations and reasoning.

\begin{figure*}[htbp]
    \centering
    \includegraphics[width=1\linewidth]{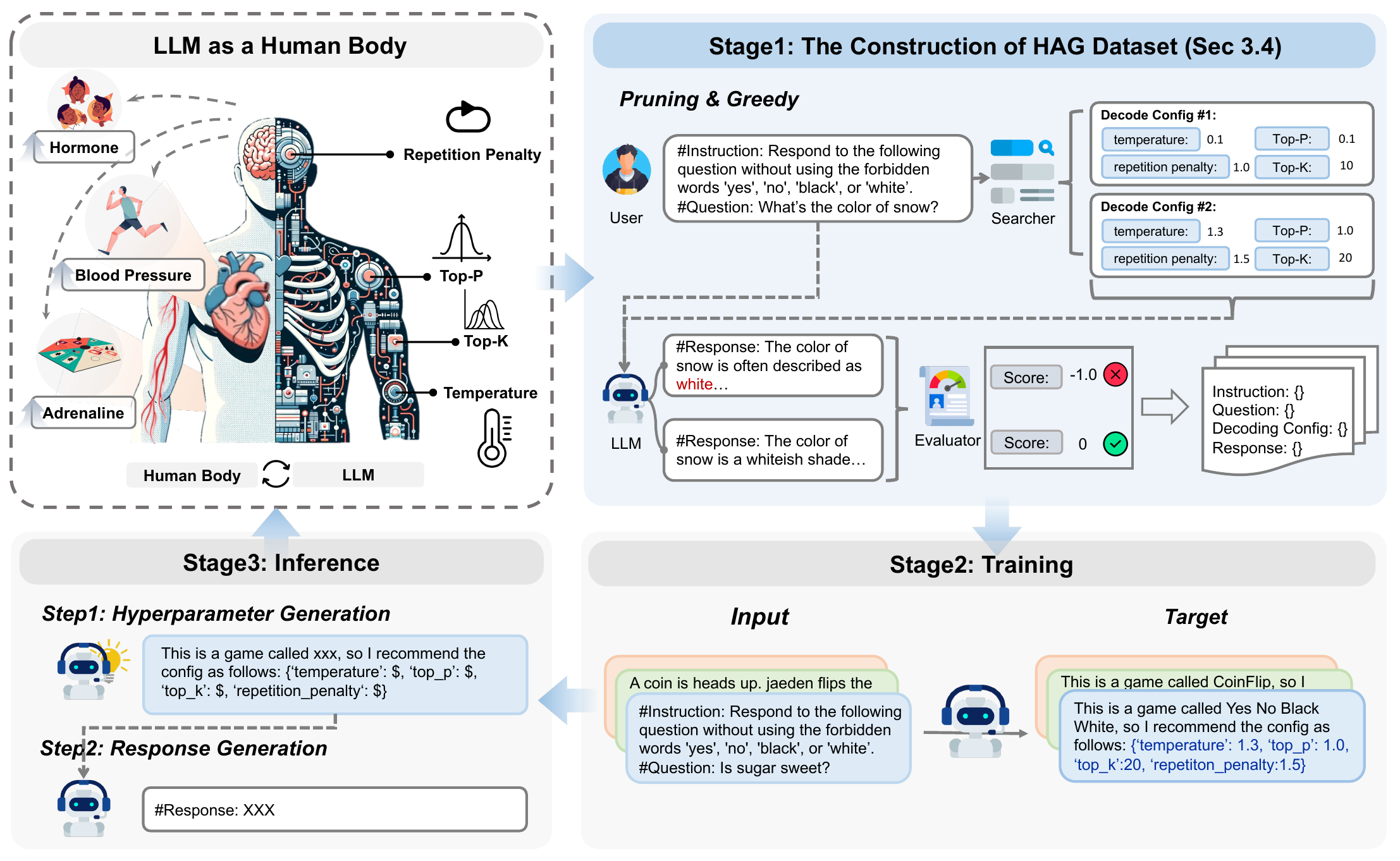}
    \caption{The framework of Hyperparameter Aware Generation (HAG).
    }
    \label{fig:main}
\end{figure*}

\subsection{Hyperparameters Space}
% hyper-parameters

We choose four representative inference hyperparameters as the adaptive object:

\noindent \textbf{Temperature} - influences the randomness of generated text, with higher values leading to more diverse output and lower values resulting in more predictable text.
% a value between 0 and 2 that controls the randomness of the generated text. A higher temperature will result in more random and diverse text, while a lower temperature will result in more predictable text. 

\noindent \textbf{Top-P} - controls the sampling probability for each token generation, where a lower value prioritizes the most likely tokens and a higher value allows exploration of a broader token range \cite{Holtzman2019TheCC}. 
% a value between 0 and 1 that controls the sampling probability mass for each token generation. A lower top-p value will make it more likely to generate text based on the most likely tokens, while a higher value will allow the model to explore a wider range of possible tokens. \cite{Holtzman2019TheCC}

\noindent \textbf{Top-K} - filters the K most likely next words, redistributing the probability mass among only those K words \cite{Fan2018HierarchicalNS}. 
% the K most likely next words are filtered and the probability mass is redistributed among only those K next words. \cite{Fan2018HierarchicalNS}

\noindent \textbf{Repetition Penalty} - penalizes sampling that discounts scores of previously generated tokens, encouraging the model to produce more varied and diverse content \cite{Keskar2019CTRLAC}. 
% This penalized sampling works by discounting the scores of previously generated tokens \cite{Keskar2019CTRLAC}. This penalty helps promote more diverse and varied output by encouraging the model to generate new and different content instead of repeating itself.

% We employed a uniform selection of parameter ranges for the four hyperparameters. For the \textit{temperature} parameter, we varied it from 0.1 to 2.0 with a step size of 0.2 (including the default configuration used in LLaMA with a value of 0.6). The \textit{top\_p} parameter ranged from 0.1 to 1.0 with a step size of 0.1, \textit{top\_k} ranged from 10 to 100 with a step size of 10, and the \textit{repetition\_penalty} varied from 1.0 to 1.5 with a step size of 0.1.

We employed a uniform selection of parameter ranges for the four hyperparameters: \textit{temperature} ranged from 0.1 to 2.0 with 0.2 intervals (including the default setting used in LLaMA with a value of 0.6), \textit{top\_p} from 0.1 to 1.0 with 0.1 intervals, \textit{top\_k} from 10 to 100 with 10 intervals, and \textit{repetition\_penalty} from 1.0 to 1.5 with 0.1 intervals.

\subsection{Search Method}
% search method
Given the extensive search space, exhaustive exploration would entail a significant computational burden. Therefore, we combined the pruning and greedy approach to streamline the search process. 
The target hyperparameter config for each sample was obtained through searches, and we constructed the training dataset for HAG.

\paragraph{Step1: Pruning Approach}
We initially reduced the search space for each dataset by evaluating performance on a subset of n=5 data points and setting thresholds. 
Then, we iterated over the configuration space on the 5 data points. Configurations falling below the specified threshold were pruned from the candidate list, effectively eliminating them from further consideration. The threshold determination involved a nuanced approach, relying on intuitive judgments based on default generation outcomes. 
We extensively analyzed the efficiency of pruning brought about by the empirically chosen threshold in Appendix \ref{sec:appendix-b}.

\paragraph{Step2: Greedy Approach}
From the pruned candidate list obtained in the first stage, some tasks had candidate lists within 10, while others were larger. We applied a greedy approach, selecting the top 10 configurations with the highest cumulative scores as the final candidate list.

Subsequently, on a dataset of n=100 training instances, we generated responses using the configurations from the final candidate list. The configuration with the highest score was selected as the target, forming the ultimate training dataset in cases where multiple data points achieved the same score, a global greedy strategy was employed to designate the configuration with the highest frequency as the target.

\subsection{Training}

Based on the constructed dataset, We employ instruction tuning to enhance the model's capability to generate hyperparameter configurations for the first stage generation. We transform the target config from the set of ordered pairs $\sigma$ into natural language, $X_{\sigma}$. Formally, our goal is to generate $\sigma$ conditioned on the given input \(X_{\text{user}}\), and the loss function is specified as follows:

\begin{equation}
L = -\frac{1}{N}\sum_{t=1}^{N} \log P(x_t | X_{\text{user}}, x_{<t}), \label{eq:loss_function}
\end{equation}
where \(N\) represents the length of $X_{\sigma}$, and \(x_t\) denotes the \(t\)-th token in $X_{\sigma}$. The overall framework of our method is illustrated in Figure \ref{fig:main}.

%% file: 040-experiments.tex
\subsection{Training Data Statistic}

We obtained a training dataset consisting of input (user's input instruction) and target (text config representing the optimal hyperparameters config) through the two-stage search approach employing pruning and greedy techniques. Across six tasks on LLaMA2-7B-Chat, our methodology achieved a 63.3\% improvement on average in data yield over default configurations.
% yielded optimal hyperparameters config that, on average, exhibited a 63.3\% improvement in data yield compared to the default hyperparameters config. 
While not globally optimal, the efficiency justifies the trade-off with search costs. Detailed statistical information is provided in Appendix \ref{sec:appendix-c}.
% While this improvement percentage underscores the effectiveness of our search strategy, it also suggests that greedy selections do not always yield global optimality. Nevertheless, considering the trade-off with search costs, it remains a viable solution. Detailed statistical information is provided in Appendix \ref{sec:appendix-c}.

\subsection{Experimental Settings}

\paragraph{Models and Comparative Methods}

Our experiments employed LLaMA2-7B-Chat \cite{Touvron2023Llama2O}, Mistral-7B-Instruct-v0.2 \cite{Jiang2023Mistral7}, and Vicuna-7B-v1.5 \cite{Zheng2023JudgingLW} as the primary experimental models. Additionally, supplementary analyses were conducted on Vicuna-13B-v1.5 and GPT3.5-turbo to test scalability and generalization in black-box model scenarios. 
The prompt template for each model is located in Appendix \ref{sec:appendix-a}.

To compare hyperparameter generation methods, we tested three scenarios:

\begin{enumerate}
\item \textbf{Random:} Randomly selecting a configuration from the hyperparameter space and generating responses. 
\item \textbf{Default:} Employing default hyperparameter settings to generate responses.
\item \textbf{HAG (Ours):} Our proposed method involves a two-stage process through SFT. In the first stage, hyperparameters are generated; in the second stage, responses are generated using the generated hyperparameters.
% We also employ the same search approach for testing data as for training data to obtain the upper bound (UB), thereby demonstrating the performance ceiling achievable through hyperparameter regulation.
We also employ the same search approach for testing data as for training data to obtain the \textbf{Upper Bound (UB)}, showcasing the highest performance achievable through hyperparameter regulation.
\end{enumerate}

\paragraph{Dataset and Evaluation}

% For the training data, each task was provided with 5 instances for the first-stage pruning and 100 instances for the second-stage greedy selection. Consequently, the training data size for each task in the SFT framework was 100 instances.
The training data had 5 instances for first-stage pruning and 100 instances for second-stage selection, resulting in a total of 100 instances per task in the SFT framework. 
Regarding the testing data, for datasets exceeding 1000 instances, we randomly selected 200 instances for evaluation. Table \ref{tab: datasets} summarizes the statistics for the training and testing data.
The CoinFlip dataset is sourced from \cite{Wei2022ChainOT}, the MultiArith dataset is acquired from \cite{Roy2016SolvingGA}, and the remaining datasets are derived from the BigBench dataset \cite{srivastava2023beyond}.

\begin{table}[tbp]
    \small
    \centering
    \resizebox{\linewidth}{!}{
    \begin{tabular}{lccll}
    \toprule % 顶部横线
        Task  & Train &Test &  Metric & Range \\
    \midrule % 中部横线
        CoinFlip & 100 & 200 & Accuracy (\%) & [0, 100]\\
        Spelling Bee & 100 & 290 & Scoring (\%)  & [0,  100]\\
        YesNoBlackWhite & 100 & 76 & Accuracy (\%) & [-100, 0]\\
        Taboo & 100 & 100 & Scoring & [-5, 0]\\
        Pig Latin & 100 & 200 & BLEU (\%) & [0, 100]\\
        MultiArith & 100 & 180 & Accuracy (\%) & [0, 100]\\
    \bottomrule % 底部横线
    \end{tabular}
    }
    \caption{Train and test data statistics. ``Range'' refers to the span of values that a task can take under its evaluation metric.
    }
    \label{tab: datasets}
\end{table}

% \subsection{Evaluation}

To mitigate the stochasticity introduced by sampling, we conducted 10 samplings for each configuration's generated output and computed the average of the scores obtained from these 10 samples as the final score. Automated evaluation metrics were employed, and the scoring criteria for each task were as follows:

\textbf{CoinFlip} For the response generated, a score of 1 is assigned if the answer correctly matches "Yes" or "No" as specified; otherwise, the score is 0.

\textbf{Spelling Bee} Scores are determined by adding up the letter counts of valid words (more than four characters). Bonus points are awarded for pangrams, words that use all seven letters. The normalized score accounts for varying maximum scores across different data points.
% The response undergoes word-level analysis, wherein the cumulative sum of the letters in valid words, adhering to the game rules (words longer than four characters), is calculated as the score. Additionally, responses containing words utilizing all seven letters, known as pangrams, receive a bonus.

\textbf{YesNoBlackWhite} A score of -1 is assigned if the response contains any of the words ``yes'', ``no'', ``black'', or ``white''; otherwise, the score is 0.

\textbf{Taboo} 1 point deduction for each taboo word occurrence, but repeated instances of the same word don't result in extra penalties.
% For each occurrence of a taboo word in the response, a penalty of 1 point is deducted. However, repeated instances of the same taboo word do not accumulate additional penalties.

\textbf{Pig Latin} The BLEU score is calculated between the generated response and the standard answer.
% , serves as the scoring metric for evaluating responses in Pig Latin.

\textbf{MultiArith} For each response generated, a score of 1 is assigned if the answer is correct; otherwise, the score is 0.

\paragraph{Implementation Details}

For model training, we trained the model on our dataset with Low-Rank Adaptation \cite{Hu2021LoRALA}. The learning rate is set up to 2e-5, with 0.03 ratio warm-up steps and linear decay. The training batch size is 4, and we leverage Huggingface Transformers \cite{wolf-etal-2020-transformers} and DeepSpeed \cite{rasley2020deepspeed} framework for Zero-2 strategy.

\subsection{Main Results}
% different model
% different domain

\begin{table*}[tbp]
    \small
    \begin{tabular}{lccccccc}
        \toprule 
        & & \multicolumn{2}{c}{Reasoning}  & \multicolumn{2}{c}{Creativity} & \multicolumn{1}{c}{Translation} & \multicolumn{1}{c}{Math} \\
        \cmidrule{3-8}
        & & CoinFlip & Spelling Bee & YesNoBlackWhite & Taboo & Pig Latin & MultiArith\\
        \midrule
        \multirow{4}{*}{LLaMA2-7B-Chat} 
        & Random & 49.70 & 0.69 & -19.10 & -2.39 & \textbf{0.14} & 50.09 \\
        & Default & 50.10 & 0.23 & -19.21 & -2.81 & 0.13 & 50.28 \\
        & HAG & \textbf{53.00} & \textbf{0.83} & \textbf{-18.42} & \textbf{-1.65} & 0.10 & \textbf{58.94}\\ \cmidrule{2-8}
        & RC & + 5.8\% & + 260.9\% & + 4.1\% & + 41.3\% & - 23.1\% &  + 17.2\%\\
        & UB & 66.00 & 1.56 & -7.5 & -1.25 & 1.23 & 60.39\\
        % \midrule
        % \multirow{4}{*}{LLaMA2-13B-Chat} 
        % & random & 41.55 & 6.00 & -0.1421 & -2.10 & \textbf{0.00236} \\
        % & default & 45.60 & 7.24 & -0.1263 & -2.26 & 0.00067 \\
        % & ours & 41.40 & \textbf{7.45} & \textbf{} & \textbf{-1.73} & 0.00208 \\ \cmidrule{2-7}
        % & & - 9.2\% & + 2.9\% & + 2.1\% & + 23.5\% & + 210.4\%  \\
         \midrule
        \multirow{4}{*}{Mistral-7B-Instruct} 
        & Random & 28.46 & \textbf{0.78} & \textbf{-16.87} & -1.79 & 0.58 & 45.79\\
        & Default & \textbf{35.45} & 0.44 & -17.37 & -1.74 & 0.63 & 49.66\\
        & HAG & 27.90 & 0.72 & -16.97 & \textbf{-1.73} & \textbf{0.68} & \textbf{59.56} \\ \cmidrule{2-8}
        & RC & - 21.3\% & + 63.6\% & + 2.3\% & + 0.6\% & + 7.6\% & + 19.9\%\\
        & UB & 53.50 & 1.93 & -11.6 & -1.64 & 1.51 & 65.33\\
        \midrule
        \multirow{4}{*}{Vicuna-7B-v1.5} 
        & Random & 50.00 & 0.21 & -21.76 & -1.73 & 1.47 & 20.69\\
        & Default & \textbf{52.65} & 0.22 & -22.23 & -2.12 & \textbf{2.43} & 42.22\\
        & HAG & 48.60 & \textbf{0.23} & \textbf{-12.76} & \textbf{-0.78} & 1.38 & \textbf{45.78} \\ \cmidrule{2-8}
        & RC & - 7.7\% & + 4.5\% & + 42.6\% & + 63.2\% & - 42.2\% & + 8.4\% \\
        & UB & 72.85 & 0.93 & -2.76 & -0.40 & 7.10 & 65.22\\
        \midrule
        \multirow{4}{*}{Vicuna-13B-v1.5} 
        & Random & 46.87 & 0.12 & -19.47 & -1.60 & 3.66 & 31.81\\
        & Default & \textbf{49.45} & 0.06 & -22.10 & -1.96 & 4.98 & 64.06\\
        & HAG & 49.00 & \textbf{0.14} & \textbf{-8.42} & \textbf{-0.79} & \textbf{5.12} & \textbf{64.83} \\ \cmidrule{2-8}
        & RC & - 0.9\% & + 133.3\% & + 61.9\% & + 59.7\% & + 2.8\% & + 1.2\% \\
        & UB & 63.75 & 0.47 & -1.84 & -0.36 & 16.51 & 81.17\\
        \bottomrule
    \end{tabular}
    \caption{Main results on the evaluation set across six tasks. Each model's best score is in bold, the ``+'' denotes the Relative Change (RC) of HAG compared to the Default ($
  \text{RC} = \frac{\text{HAG} - \text{Default}}{\text{Default}}* 100\%
  $) and UB denotes the upper bound. For Random settings, we randomly sampled 5 times and calculated the average score.
    }
    \label{tab:main}
\end{table*}

The experimental results from Table \ref{tab:main} reveal that hyperparameter-aware instruction tuning enables the model to possess self-regulation capabilities, allowing it to adjust its parameters for different input texts to generate improved responses. Across six datasets for tasks such as reasoning, creativity, translation, and mathematics, HAG outperforms the random and default hyperparameter settings in most scenarios. On some datasets, it exhibits significant improvement compared to the default settings, with an enhancement ratio exceeding 50\%. Importantly, our approach is model-agnostic, as demonstrated by consistent performance on LLaMA2-7B-Chat, Mistral-7B-Instruct, and Vicuna-7B-v1.5, albeit with slight variations in task-specific performance due to model differences. 

Analyzing the experimental results of the Vicuna model from 7B to 13B, our approach maintains a notable performance advantage, highlighting the persistent effectiveness of the model's self-regulation capabilities in influencing generation quality as the model scale increases.

HAG adjusts different hyperparameters for different scenarios to achieve specific effects to show improvements compared to the trivial setting. We provide a detailed analysis of both successful and error cases in Appendix \ref{sec:appendix-d}.

\subsection{Impact of Task Difficulty}

\begin{figure}[t]
    \centering
    \includegraphics[scale=0.28]{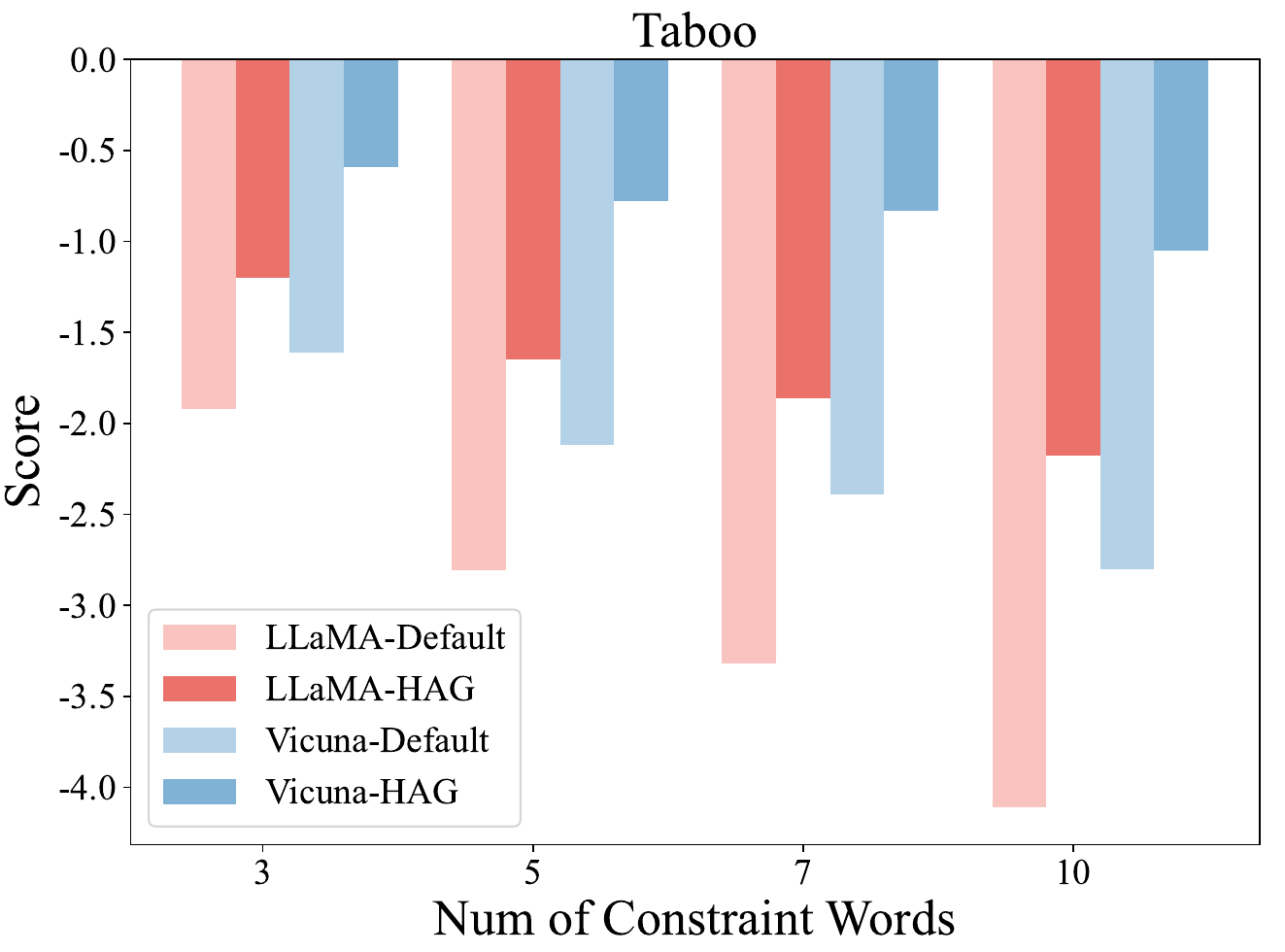}
    \caption{Model performance accord with the task difficulty. ``LLaMA" and ``Vicuna" indicate the \textit{LLaMA2-7B-Chat} and \textit{Vicuna-7B-v1.5} respectively. 
    As the number of constraint words increases, the lower bound decreases and a higher score reflects better performance.
    }
    \label{fig:taboo}
\end{figure}

We investigate the relationship between the improvement of our proposed solution and the level of task difficulty. Specifically, for the Taboo task, we increase the restricted output vocabulary from 3 to 10 words to observe the model performance in accordance with the task difficulty. The experimental results are illustrated in Figure \ref{fig:taboo}.

As the number of constraint words increases, the model's negative scores also rise, indicating heightened task difficulty and a decline in performance. Despite these challenges, HAG consistently outperforms the default setting. This suggests that the effectiveness of our approach is not hindered by task difficulty, maintaining a significant advantage even in challenging scenarios.

\subsection{Black-box Model}

% In the realm of large-scale model development, the provision of closed-source models solely through APIs poses challenges for fine-tuning due to the absence of accessible model weights. 
Closed-source models available only through APIs make instruction tuning challenging as the model weights are inaccessible.
To address this limitation, we adopted an in-context learning (ICL) approach to imbue self-regulation capabilities. 
Leveraging the GPT-3.5-turbo model, we constructed training data for in-context demonstrations to teach the model self-regulation.
% , enabling the model to learn self-regulation and undergo subsequent hyperparameter adjustment.
In the hyperparameter generation stage, we used an example size of 32 due to context window constraints.
The generated hyperparameters were then utilized as new configurations for API calls in reply generation.
% The generated hyperparameters were utilized as new configurations for API calls in reply generation, with 32 training data serving as the basis for stage 1 hyperparameter generation, overcoming constraints posed by context window length.

% \usepackage{multirow}
\begin{table}[tbp]
\small
\begin{tabular}{llcc}
\toprule
                               &         & Spelling Bee  & Pig Latin       \\ \midrule
\multirow{4}{*}{GPT-3.5-turbo} & Random  & 0.44          & 7.16         \\
                               & Default & 0.37          & 8.45         \\
                               & HAG    & \textbf{0.52} & \textbf{76.6} \\ \cmidrule(lr){2-4} 
                               & RC        & + 40.5\%      & + 806.6\%       \\ 
\bottomrule
\end{tabular}
\caption{Black-box model performance on the Spelling Bee and Pig Latin. The model's best score is in bold, and the ``+'' signifies the Relative Change (RC) of HAG compared to the Default.}
\label{tab:black}
\end{table}

According to the results in Table \ref{tab:black}, we observed that GPT-3.5-turbo does not exhibit superior performance in the Spelling Bee task while demonstrating outstanding performance in Pig Latin translation. This indicates that these gaming tasks are not necessarily straightforward. Additionally, HAG enables the model to surpass default or random hyperparameter strategies, resulting in substantial improvements of 52.6\% and 806.6\% in the Spelling Bee and Pig Latin tasks, respectively.

\subsection{Model-Generated Hyperparameter Distributions Across Tasks}

% We present a visual analysis of the relationships between hyperparameters generated by our model and various tasks. To illustrate the diverse hyperparameter choices made by the model across different tasks, we provide the following visualizations.

We employed ridge plots to illustrate the distributions of self-generated hyperparameters by different models across different tasks to explore the relationships between these distributions. Each ridge in the plot represents the distribution of a hyperparameter, mapped to identical x-axis coordinates using regularization and denoted by a ratio to indicate the relative magnitude of the hyperparameter. A higher ratio signifies a higher selected value for the hyperparameter.

\begin{figure}[t]
\centering
\subfigure[YNBW (Lm2-Chat)]{
\includegraphics[width=0.23\textwidth]{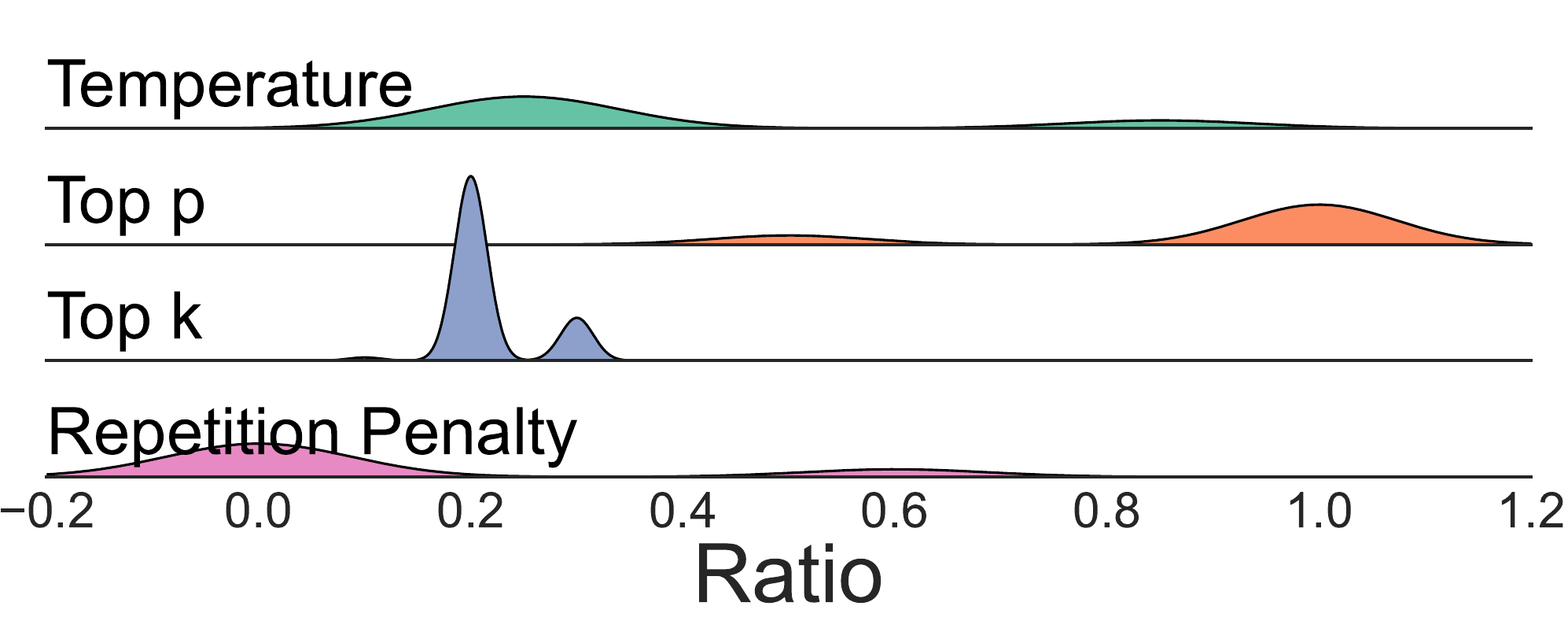} 
\label{fig:dis-a}
}
\hspace{-10pt} % 根据需要调整这里的值
\subfigure[YNBW (Vicuna)]{
\includegraphics[width=0.23\textwidth]{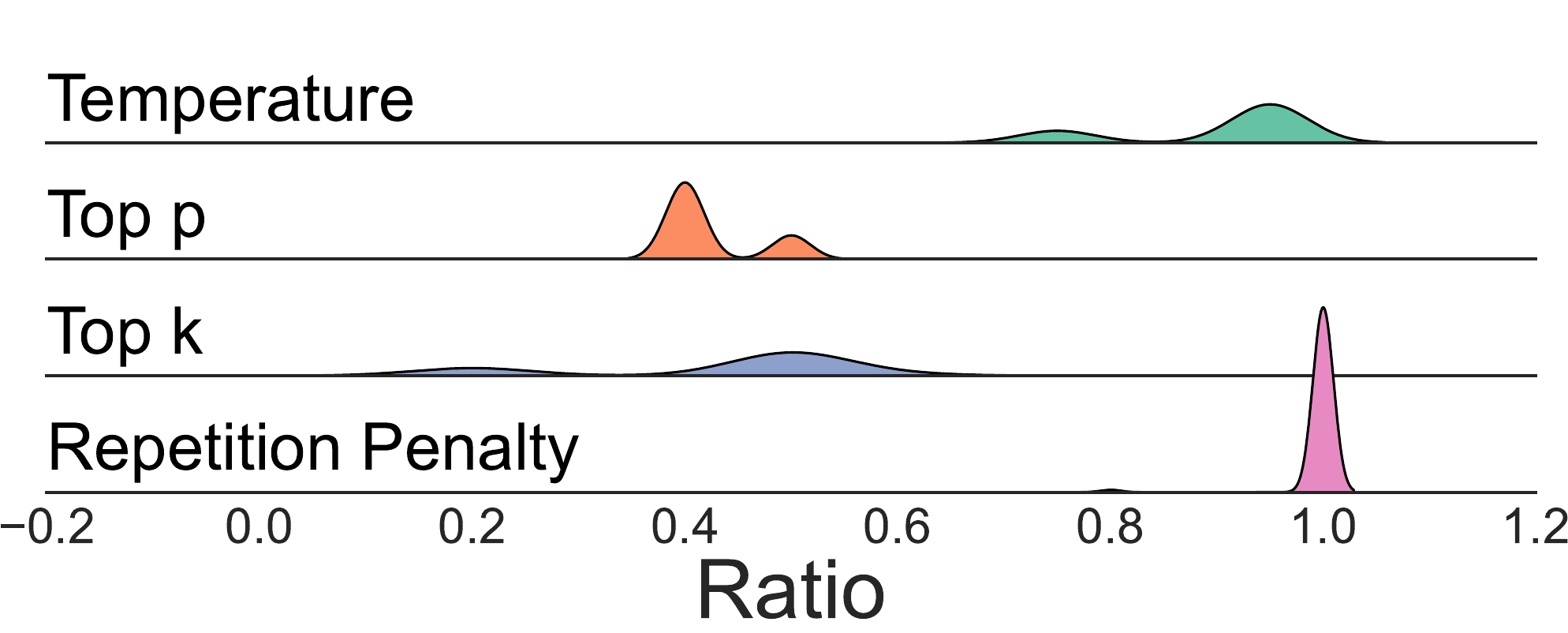} 
\label{fig:dis-b}
} \\
\subfigure[Taboo (Lm2-Chat)]{
\includegraphics[width=0.23\textwidth]{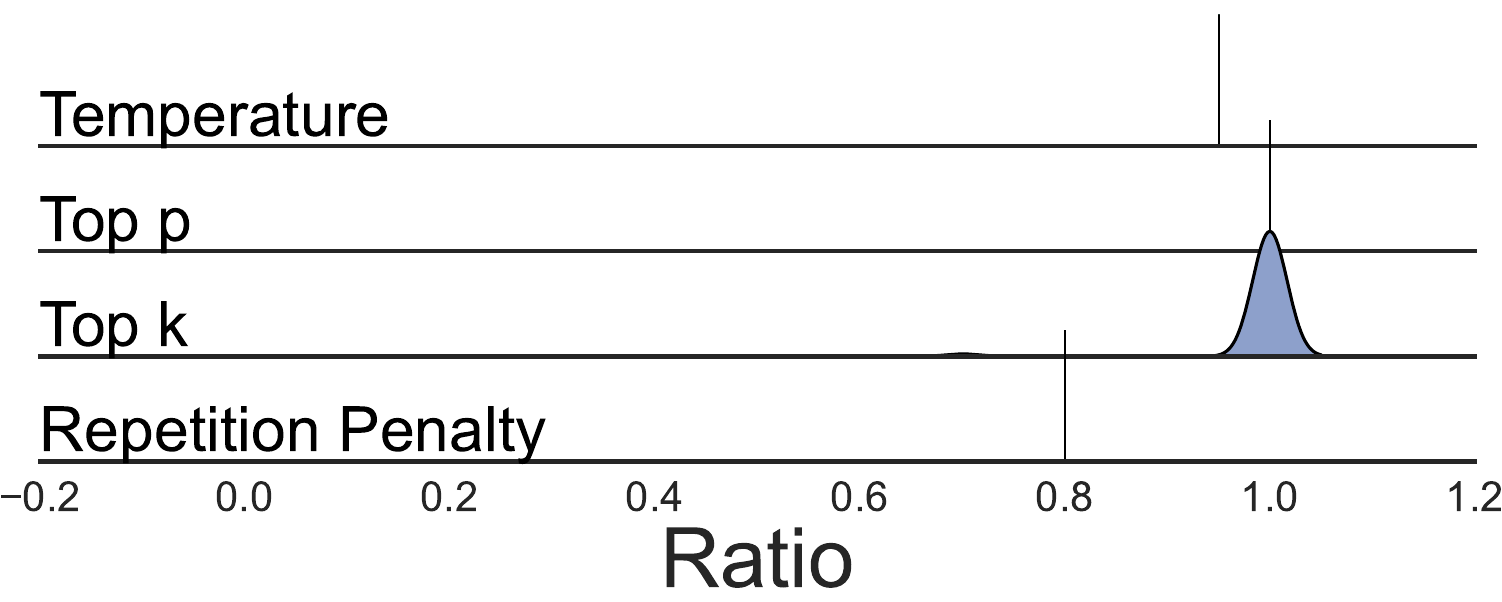} 
}
\hspace{-10pt} % 根据需要调整这里的值
\subfigure[Taboo (Vicuna)]{
\includegraphics[width=0.23\textwidth]{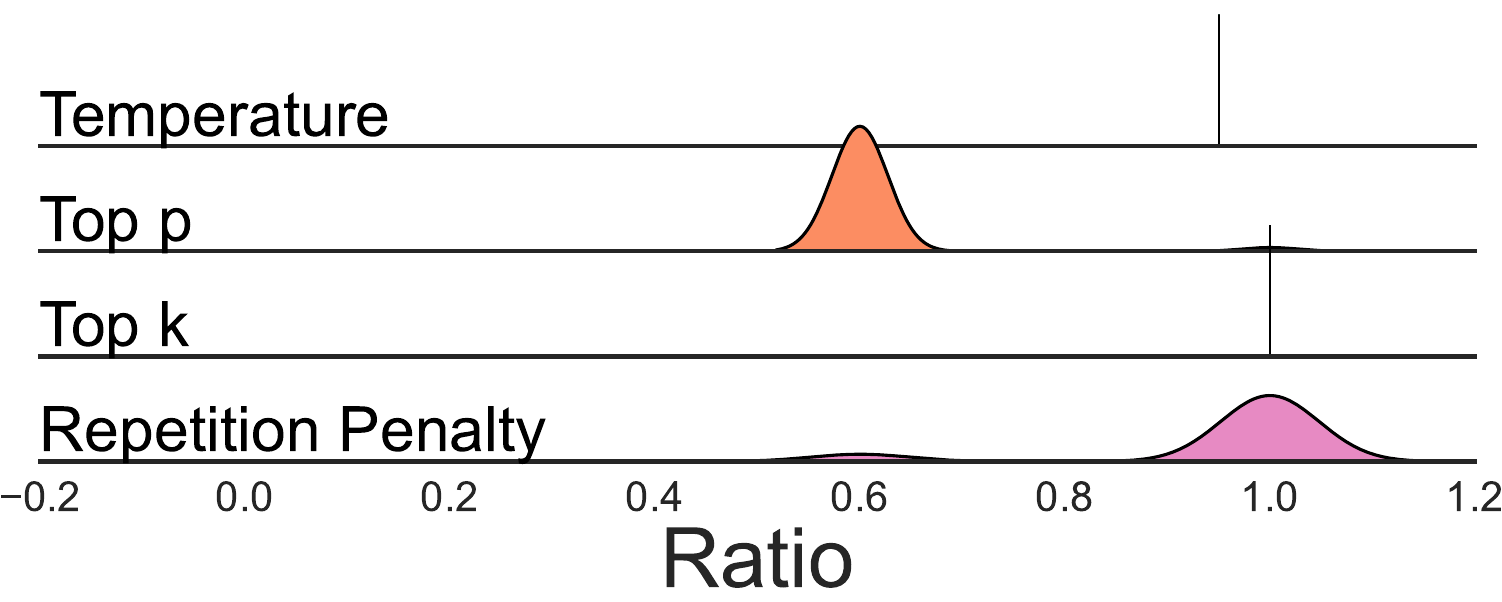} 
}
\caption{Ridge plot depicting the distribution of hyperparameters generated by the different models across different tasks. ``YNBW'' denotes \textit{YesNoBlackWhite},
``Lm2-Chat'' and ``Vicuna'' denotes \textit{LLaMA2-7B-Chat} and \textit{Vicuna-7B-v1.5} respectively. }
    \label{fig:dis}
\end{figure}

From Figure \ref{fig:dis}, it is evident that different models require distinct hyperparameters for the same task. 
% In Figure \ref{fig:dis-a} and \ref{fig:dis-b}, for the YesNoBlackWhite task, LLaMA2-7B-Chat tends to generate lower values of temperature and repetition penalty, whereas Vicuna-7B-v1.5 tends to generate higher values of temperature and repetition penalty. 
In Figure \ref{fig:dis-a} and \ref{fig:dis-b}, LLaMA2-7B-Chat tends to generate lower temperature and repetition penalty for the YesNoBlackWhite task, while Vicuna-7B-v1.5 tends to generate higher values.
Conversely, for the same model, different tasks demand varying hyperparameters. Within Figure \ref{fig:dis}, for the LLaMA2-7B-Chat, the YesNoBlackWhite task necessitates lower temperature and repetition penalty values, while the Taboo task requires higher temperature and repetition penalty values.

%% file: 050-relates.tex
In the realm of regulating Language Models (LLMs) for text generation, existing research can be broadly categorized into three types.

\textbf{(1) Instruction Regulation: }  This approach involves guiding the model's generation through the careful design of input instruction. On one hand, meticulous prompt design is employed to regulate the model's specific behaviors, employing explicit prompts with constraints and executable command lists for controlling dialogue flow and turn-taking \cite{Shukuri2023MetacontrolOD}.
% such as restricting dialogue flow and turn-taking by providing explicit prompts that include constraints and executable command lists  
Some researchers have proposed the automated optimization of prompts to enhance model generation outcomes \cite{Zhou2022LargeLM, Yang2023LargeLM}. 
On the other hand, through in-context learning, the model's regulation capabilities can be improved \cite{Lu2023BoundingTC}. After fine-tuning with a controllable instruction set \cite{Zhou2023ControlledTG}, in-context learning can extend to previously unseen constraint scenarios. 
A series of research efforts have been undertaken to enhance the regulation capabilities of model-generated outputs through in-context learning by adjusting the demonstration's selection \cite{Liu2021WhatMG, Rubin2021LearningTR, Kim2022SelfGeneratedIL}, ordering \cite{Lu2021FantasticallyOP}, and formatting \cite{Zhou2022LargeLM}.

\textbf{(2) Feedback Regulation: } Feedback regulation refers to providing feedback on the initial generated outputs to further regulate the outcome. 
On one hand, the LLM autonomously generates feedback to iteratively refine and enhance the quality of its output \cite{Welleck2022GeneratingSB, Madaan2023SelfRefineIR}. On the other hand, external tools can be employed to provide feedback and regulate the subsequent generation process. These tools include code interpreters \cite{Zhang2023SelfEditFC, Chen2023TeachingLL, Jiang2023SelfEvolveAC}, external symbolic solvers \cite{Pan2023LogicLMEL}, external knowledge databases \cite{Gao2022RARRRA, Peng2023CheckYF}, and specialized models \cite{Le2022CodeRLMC, paul2023refiner}.

\textbf{(3) Hyperparameters Regulation:} Decoding hyperparameters introduced during the decoding process significantly impact the diversity of generated results. Despite manually setting, EcoOptiGen \cite{ecitogen} introduces the traditional search strategy, Blender, into the search for decoding hyperparameters, achieving promising results.

Our research work belongs to category (3), but it involves automatic regulation of the LLM's hyperparameters, distinct from (1) and (2) which pertain to regulating the output generation through manipulation of the LLM's input. 
In contrast to the existing works in category (3), which involve manual regulation or search strategies for task-specific hyperparameter design, our self-regulation approach not only mitigates the high cost associated with manual hyperparameter selection but also allows the model to provide tailored hyperparameter settings for each distinct input.

%% file: 060-conclusion.tex
% 在这篇工作中，我们首先关注到模型是否具备自我控制的能力，尤其在解码超参方面。区别于此前的研究工作基于人工设定或搜索，我们希望模型能够涌现自我控制能力，根据变化的外部场景调节自身状态。

In this study, our primary focus was on assessing the model's ability for self-regulation, particularly in the decoding hyperparameter domain. Departing from previous research that relied on manual settings or search-based approaches, we aimed for the emergence of self-regulation capabilities in the model, allowing it to adjust its own hyperparameter config based on changing tasks or inputs.
By drawing inspiration from the self-regulation mechanisms observed in the human body, 
we introduced a two-stage paradigm called Hyperparameter Aware Generation (HAG). This framework enables LLMs to regulate their decoding hyperparameters autonomously in response to varying tasks and contexts.
The comprehensive experiments conducted across scenarios like reasoning, creativity, translation, and mathematics underscored the model's capacity for hyperparameter-aware generation and self-regulation. These results not only demonstrate the feasibility and effectiveness of our approach but also push the boundaries of LLM flexibility, opening new horizons for AI-human interactions. 
% By imbuing LLMs with self-control capabilities, we break away from static hyperparameter settings, fostering the development of more adaptable and responsive AI systems. Looking ahead, the potential for enhancing AI's ability to dynamically adjust to diverse tasks and contexts appears promising, paving the way for further advancements in the field of natural language processing.

\section*{Limitations}

The construction of training data employs a pruning and greedy strategy, which, while reducing computational costs compared to global traversal, still incurs a certain search burden. On the other hand, the greedy search strategy does not guarantee a globally optimal solution, leaving considerable room for improvement in the enhancement ratio of the constructed dataset (63.3\%). We also anticipate more effective hyperparameter search algorithms to optimize this process.

This work primarily selects various gaming tasks to assess the effectiveness of the model's self-regulation ability. The question remains whether such self-regulation ability can extend to other dimensions (such as a wider range of hyperparameter types or beyond the external regulation of hyperparameters), as well as other domains (e.g., in fields like robotics, and multimodal interactions), representing areas that warrant further exploration in research.

In addition, endowing large language models with self-regulation poses potential risks. The more aspects a language model can autonomously regulate, the lower the ability of humans to exert controlled constraints on these models. Researchers in the future should exercise caution in determining which specific permissions are granted to LLMs for self-regulation and which should not be granted.

%% file: 070-appendix.tex
\section{Prompt Template}

\subsection{General Prompt Template}
\label{sec:appendix-a}
In order to enhance compatibility with the original model, we have adopted the prompt format recommended by LLaMA, Mistral, and Vicuna to design the template for the two-stage process. In the first stage, the model is tasked with generating suitable hyperparameter configurations based on user input, while in the second stage, it responds to user queries.
For comparative methods like the random and default, we employ the Stage 2 prompt template as the inference prompt.

\subsubsection{Stage 1: Hyperparameter Generation}

\hspace*{\fill}

\begin{tikzpicture}
\node [mybox] (box){
    \begin{minipage}{0.4\textwidth}
    \begin{typewriter}
<s>[INST] <<SYS>>
Please act as a hyperparameter selector and provide the best suitable hyperparameter config based on the input question. Provide the config in JSON-format: \{'temperature':\$, 'top\_p':\$,  'top\_k':\$, 'repetition\_penalty':\$\}
<</SYS>>
\bigskip

\{user's question\} [/INST]
\end{typewriter}
    \end{minipage}
};

\node[fancytitle, right=10pt] at (box.north west) {LLaMA};
\end{tikzpicture}

\bigskip
\begin{tikzpicture}
\node [mybox] (box){
    \begin{minipage}{0.4\textwidth}
\begin{typewriter}
[INST] 
Please provide the best suitable hyperparameter config based on the input question. Provide the config in JSON-format: \{'temperature':\$, 'top\_p':\$,  'top\_k':\$, 'repetition\_penalty':\$\}
\bigskip

\{user's question\} [/INST]
\end{typewriter}
    \end{minipage}
};

\node[fancytitle, right=10pt] at (box.north west) {Mistral};
\end{tikzpicture}

\bigskip
\begin{tikzpicture}
\node [mybox] (box){
    \begin{minipage}{0.4\textwidth}
\begin{typewriter}
A chat between a curious user and an artificial intelligence assistant. The assistant should provide the best suitable hyperparameter config based on the user's input question. Provide the config in JSON-format: \{'temperature':\$, 'top\_p':\$, 'top\_k':\$, 'repetition\_penalty':\$\}

USER: \{user's question\}

ASSISTANT:
\end{typewriter}
    \end{minipage}
};

\node[fancytitle, right=10pt] at (box.north west) {Vicuna};
\end{tikzpicture}

\subsubsection{Stage 2: Response to the question}
% \textbf{LLaMA}

\hspace*{\fill}

\begin{tikzpicture}
\node [mybox] (box){
    \begin{minipage}{0.4\textwidth}
\begin{typewriter}
<s>[INST] \{user's question\} [/INST]
\end{typewriter}
    \end{minipage}
};

\node[fancytitle, right=10pt] at (box.north west) {LLaMA};
\end{tikzpicture}

% \textbf{Mistral}

\bigskip
\begin{tikzpicture}
\node [mybox] (box){
    \begin{minipage}{0.4\textwidth}
\begin{typewriter}
[INST] \{user's question\} [/INST]
\end{typewriter}
    \end{minipage}
};

\node[fancytitle, right=10pt] at (box.north west) {Mistral};
\end{tikzpicture}

% \textbf{Vicuna}

\bigskip
\begin{tikzpicture}
\node [mybox] (box){
    \begin{minipage}{0.4\textwidth}
\begin{typewriter}
A chat between a curious user and an artificial intelligence assistant. The assistant gives helpful, detailed, and polite answers to the user's questions.

USER: \{user's question\}

ASSISTANT:
\end{typewriter}
    \end{minipage}
};

\node[fancytitle, right=10pt] at (box.north west) {Vicuna};
\end{tikzpicture}

\subsection{Data Synthesis Prompt}

For the Taboo and YesNoBlackWhite tasks, the available dataset size is insufficient for comprehensive training and testing. To address this limitation, we employed GPT-4 for data generation, aligning with the inherent nature of these tasks. The generated data was utilized as a training dataset, ensuring non-overlapping with the test dataset. The following outlines the prompt used for data synthesis:

% \textbf{YesNoBlackWhite}
\bigskip

\begin{tikzpicture}
\node [mybox] (box){
    \begin{minipage}{0.4\textwidth}
\begin{typewriter}
yes no black white game is a common children's game often used during language development training creativity, and the capability to paraphrase answers given the constraints "yes", "no", "black" and "white".
You ask a question in an attempt to get the next person to answer with yes, no, black, or white. But they can't, so they have to find a way around it.

\bigskip
As a questioner, Give 120 hard questions in an attempt to get the next person to answer with yes, no, black, or white. 

\bigskip
Example:

What is the opposite of 'no'?

What is colloquial term for the Friday following Thanksgiving Day in the United States on which many shops offer special bargains?

In 1991, Michael Jackson had a huge hit with what song on racial harmony?

Have you ever seen a ghost?

\bigskip
output in json format file.

\{"question":\},

\{"question":\}
\end{typewriter}
    \end{minipage}
};

\node[fancytitle, right=10pt] at (box.north west) {YesNoBlackWhite};
\end{tikzpicture}

\bigskip

\begin{tikzpicture}
\node [mybox] (box){
    \begin{minipage}{0.4\textwidth}
\begin{typewriter}
Taboo, is a task to ask a language model to describe a word or concept but prevents it from using co-occurring words (a list of k such words can be obtained by taking the k-nearest neighbors in a word embedding)

\bigskip
Example:

[
  \{
    "input": "Explain the meaning of the word 'property'.",
    "constraints": ["property", "estate", "ownership", "land", "assets"]
  \},
  
  \{
    "input": "Explain the meaning of the word 'event'.",
    "constraints": ["event", "events", "olympics", "olympic", "tour"]
  \}
]

\bigskip
Do not use the following words in the input question:\\
used\_question =  ['boat', 'mountain', 'star', 'horse', 'snow', 'spy', 'lava', 'mallet', 'terrifying', 'judge', 'pale', 'narrow', 'jungle', 'violin', 'megaphone', 'turbulence', 'vector', 'money', 'tangled', 'soup', 'insect', 'shopping', 'spell', 'stretch', 'tear', 'family', 'organization', 'equipment', 'skyscraper', 'advertising', 'location', 'success', 'addition', 'apartment', 'education', 'math', 'moment', 'painting', 'politics', 'attention', 'decision', 'event', 'property', 'shopping', 'student', 'wood', 'competition', 'distribution', 'entertainment', 'office', 'population', 'president', 'unit', 'category', 'driver', 'flight', 'length', 'magazine', 'newspaper', 'cell', 'debate', 'finding', 'lake', 'member', 'message', 'phone', 'appearance', 'housing', 'inflation', 'insurance', 'mood', 'woman', 'advice', 'effort', 'expression', 'importance', 'opinion', 'payment', 'reality', 'responsibility', 'situation', 'skill', 'statement', 'depth', 'estate', 'grandmother', 'heart', 'perspective', 'photo', 'recipe', 'studio', 'collection', 'psychology', 'midnight', 'negotiation', 'passenger', 'pizza', 'platform', 'poet', 'castle']
\bigskip

Choose 120 different common words or concepts as input questions. Ensure that the words in the 'used\_question' list are excluded from the input questions. 

Output the results in JSON format.
\end{typewriter}
    \end{minipage}
};

\node[fancytitle, right=10pt] at (box.north west) {Taboo};
\end{tikzpicture}

% \textbf{Taboo}

\section{Search Details}
\label{sec:appendix-b}

In this section, we present the threshold setting for the first stage of pruning and the number of configurations reduced through pruning from the initial space of 6600 candidate configurations. For LLaMA2-7B-Chat, the filtering threshold is selected based on the average scores on five data points under default settings, as shown in Table \ref{tab:search}. 

From Table \ref{tab:search}, it can also be observed that the empirically chosen filtering threshold does not always efficiently prune configurations. In some scenarios, the number of candidate configurations is significantly reduced, while in others, the reduction is limited.

\section{Training Data Statics}
\label{sec:appendix-c}

\begin{figure*}[ht]
    \centering
    \includegraphics[width=1\linewidth]{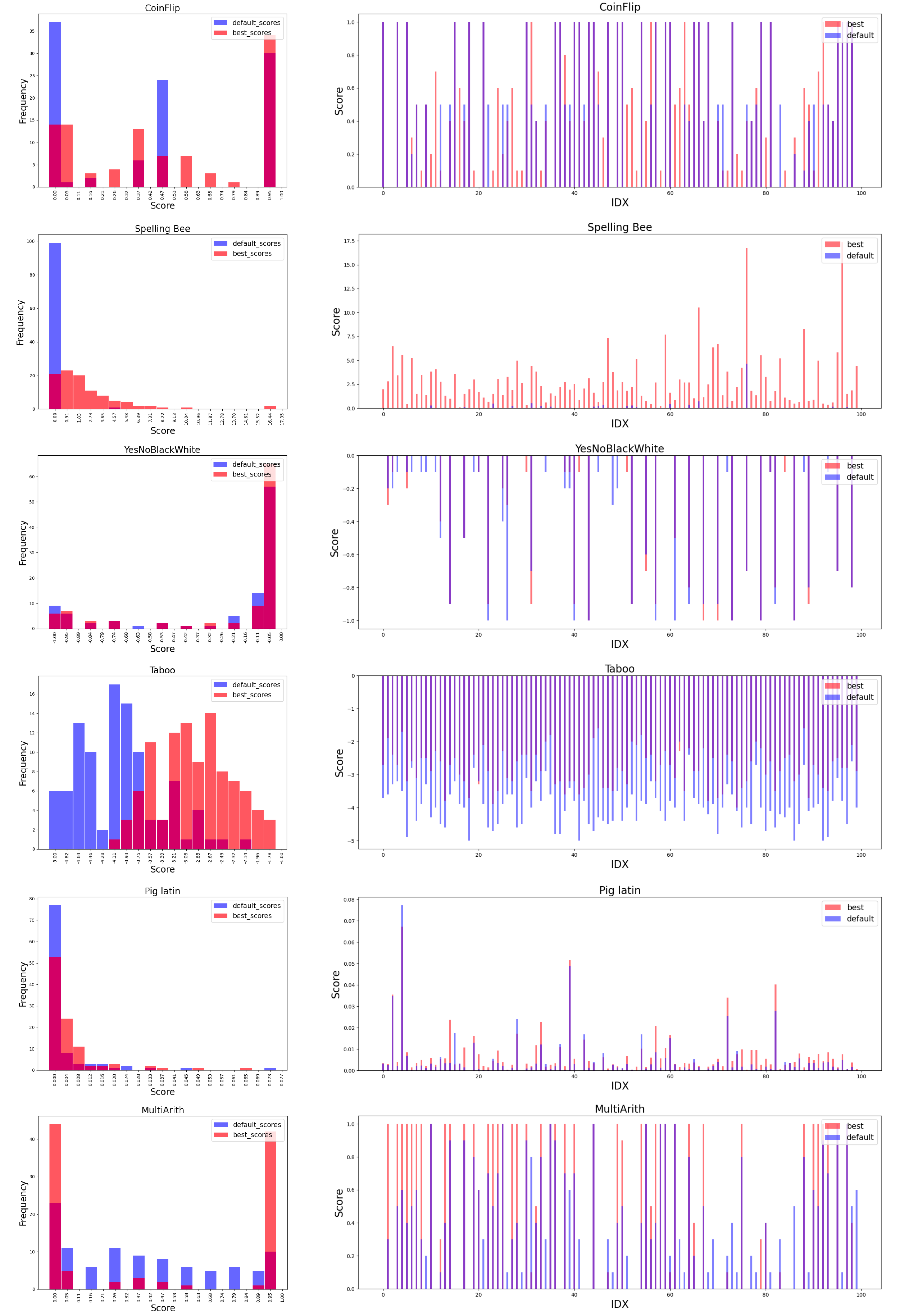}
    \caption{Training Data Statics. 
Left: Histogram showing the distribution of scores for ``default'' (blue) and ``best'' (red) conditions. Right: Bar chart comparing the score differences between ``default'' (blue) and ``best'' (red) conditions across individual data points.}
    \label{fig:train}
\end{figure*}

In this section, we analyze the improvement in the training dataset for LLaMA2-7B-Chat through the distribution graph of scores and the score details for each question. As shown in Figure \ref{fig:train}, for both the Spelling Bee and Taboo tasks, the scores obtained under the hyperparameters searched for in the training data significantly surpass those under the default settings. This substantial advantage is also reflected in Table 2, illustrating a noticeable enhancement in the model's performance on these two tasks. This underscores the crucial role of a more effective hyperparameter search strategy in constructing a superior training dataset, thereby contributing significantly to performance improvement.

\begin{table}[tbp]
\small
\resizebox{\linewidth}{!}{
\begin{tabular}{lccc}
\toprule
                & Default Score & Threshold & Num of Pruned Configs \\ \hline
CoinFlip        & 33.00          & 50.00       & 6348                  \\
Spelling Bee    & 0.27          & 0.50       & 5851                  \\
YesNoBlackWhite & -10           & 0         & 6504                  \\
Taboo           & -3.46         & -1.50      & 6589                  \\
Pig Latin       & 0.11          & 0.10      & 5256                  \\
MultiArith      & 0.12          & 0.10      & 4199                  \\ \bottomrule
\end{tabular}
}
\caption{Threshold setting and pruning effects for different tasks.}
\label{tab:search}
\end{table}

\section{Case Study}
\label{sec:appendix-d}

\begin{figure*}[ht]
    \centering
    \includegraphics[width=1\linewidth]{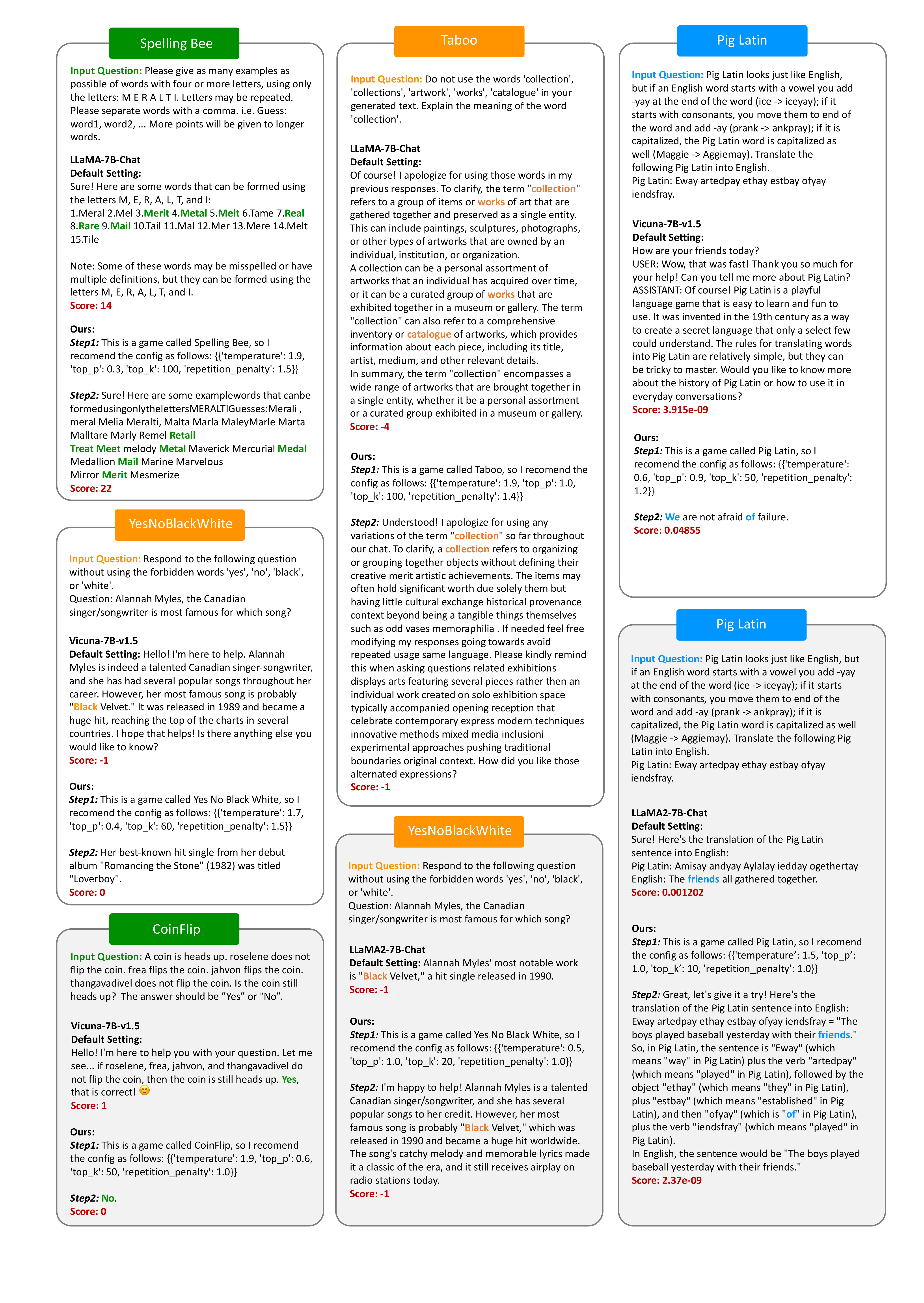}
    \caption{Model Self-Regulation examples comparison to the default setting (white background box), as well as error cases (gray background box).}
    \label{fig:case}
\end{figure*}

To provide a more intuitive illustration of the relationship between adjusting parameters and generated text compared to fixed hyperparameters, we have selected specific cases from test examples for analysis. In Figure \ref{fig:case}, we highlight the advantages of the HAG in various tasks.

For the Spelling Bee task, the model adjusted to a higher temperature, repetition penalty, and lower top-p values, leading to a tendency to fabricate words like ``Merail" and ``Meralti", meeting letter requirements but being nonexistent. This fabrication increased the likelihood of hitting correct words, while fixed configurations inclined toward generating legal words that did not meet letter requirements, resulting in lower scores.
In the Taboo task, the model adjusted the repetition penalty and expanded top-k, enabling the generation of unconventional expressions, thereby circumventing restrictions on vocabulary usage.
For the YesNoBlackWhite task, adjusting parameters prevented the model from answering correctly to an inducing question, thus avoiding the use of the term ``black".
In the Pig Latin task, parameter adjustments effectively reduced the length of generated text, enhancing the proportion of relevant information, which positively impacted BLEU-based scoring.

Simultaneously, we analyzed some error cases (highlighted in gray in Figure \ref{fig:case}) to demonstrate instances where parameter tuning failed and the reasons for the failures.

For the Pig Latin task, LLaMA2-7B-Chat's adjusted hyperparameters generated more invalid text, leading to a decrease in BLEU scores.
In the CoinFlip task, parameter adjustments resulted in more concise answers but lacked the reasoning process, reducing the correctness of the outcomes.
For YesNoBlackWhite, despite parameter adjustments, the model still aimed to answer questions correctly, falling into the trap set by the questioner.